\documentclass{llncs}
\usepackage{makeidx}  
\usepackage{cite}
\usepackage{amsmath,amssymb,amsfonts}
\usepackage{algorithmic}
\usepackage{graphicx}
\usepackage{textcomp}
\usepackage{xcolor}

\begin{document}
\frontmatter          
\pagestyle{headings}  
\addtocmark{Hamiltonian Mechanics} 

%
\mainmatter              
\title{Combining Individual and Joint Networking Behavior for Intelligent IoT Analytics\vspace{-5mm}\thanks{ 
\scriptsize This work was completed during J. Vikranth’s summer
internship in 2019 at Arm Research. Prof. M. Srivastava and J. Vikranth are partially supported by the CONIX Research Center, one of six centers
in JUMP, a Semiconductor Research Corporation (SRC) program sponsored
by DARPA.}}
\author{Jeya Vikranth Jeyakumar$^{1,4}$, Ludmila Cherkasova$^1$, Saina Lajevardi$^2$, \hspace{10mm} Moray Allan$^3$, Yue Zhao$^2$, John Fry$^2$, Mani Srivastava$^4$}
\institute{Arm Research, San Jose, USA \hspace{2mm} $^2$Arm Inc, San Jose, USA \hspace{2mm}  $^3$Arm Inc, Glasgow, UK 
$^4$University of California, Los Angeles, CA, USA\vspace{-5mm}}

\titlerunning{IoTelligent}  
%
%
\authorrunning{J.Vikranth et al.} 
%

\tocauthor{Ivar Ekeland, Roger Temam, Jeffrey Dean, David Grove, Craig Chambers, Kim B. Bruce, and Elisa Bertino}

%

\maketitle              

\makeatother
\begin{abstract}
{
The IoT vision of a trillion connected devices over the next decade requires reliable end-to-end connectivity and automated device management platforms. While we have seen successful efforts for maintaining small IoT testbeds,  there are multiple challenges for the efficient management of large-scale device deployments. With Industrial IoT, incorporating millions of devices, traditional management methods do not scale well. In this work, we address these challenges by designing a set of novel machine learning techniques, which form a foundation of a new tool, {\it IoTelligent}, for IoT device management, using traffic characteristics obtained at the network level. 
The design of our tool is driven by the analysis of 1-year long networking data, collected from 350 companies with IoT deployments. The exploratory analysis of this data reveals that IoT environments follow the famous Pareto principle, such as: (i) 10\% of the companies in the dataset contribute to 90\% of the entire traffic; (ii) 7\% of all the companies in the set own 90\% of all the devices. We designed and evaluated CNN, LSTM, and Convolutional LSTM models for demand forecasting,  with a conclusion of the Convolutional LSTM model being the best. However, maintaining and updating individual company models is expensive. In this work, we design a novel, scalable approach, where a general demand forecasting model is built using the combined data of all the companies with a normalization factor. Moreover, we introduce a novel technique for device management, based on autoencoders. They automatically extract relevant device features to identify device groups with similar behavior to flag anomalous devices.
\vspace{-4mm}
}

\end{abstract}
\section{Introduction}

The high-tech industry expects a trillion new IoT devices will be produced between now and 2035~\cite{masa-1T,hima-mbed-2018, 1T}.  These devices could range from simple sensors in everyday objects to complex devices, defined by the industrial and manufacturing processes. The Internet of Things ecosystem should include the necessary components that enable businesses, governments, and consumers to seamlessly connect to their IoT devices. 
This vision requires reliable end-to-end connectivity and device management platform, which makes it easier for device owners to access their IoT data and exploiting the opportunity to derive real business value from this data.  The benefits of leveraging this data are greater business efficiencies, faster time to market, cost savings, and new revenue streams. Embracing these benefits ultimately comes down to ensuring the data is secure and readily accessible for meaningful insights.

The Arm Mbed IoT Device Management Platform~\cite{pelion-devices} addresses these requirements by enabling organizations to securely develop, provision and manage connected devices at scale and by enabling the connectivity management~\cite{pelion-connectivity} of every device regardless of its location or network type. The designed platform supports the physical connectivity across all major wireless protocols  (such as cellular, LoRa, Satellite, etc.) that can be managed through a single user interface. Seamlessly connecting all IoT devices is important in ensuring their data is accessible at the appropriate time and cost across any use case. While we could see successful examples of deploying and maintaining small IoT testbeds,  {\it there are multiple challenges in designing an efficient management platform for large-scale device deployments}. The operators of IoT environments may not be fully aware of their IoT assets, let alone whether each IoT device is functioning and connected properly, and whether enough networking resources and bandwidth allocated to support the performance objectives of their IoT networks. With the IoT devices being projected to scale to billions, the traditional (customized or manual) methods of device and IoT networks management do not scale to meet the required performance objectives. 

In this work, we aim to address these challenges by designing a set of novel machine learning techniques, which form a foundation of a new tool, {\it IoTelligent}, for IoT networks and device management, using traffic characteristics obtained at the network level.
One of the main objectives of {\it IoTelligent} is to build effective demand forecasting methods for owners of IoT ecosystems to manage trends, predict performance, and detect failures. The insights and prediction results of the tool will be of interest to the operators of IoT environments. 

For designing the tool and appropriate techniques, we utilize the unique set of real (anonymized) data, which were provided to us by our business partners. This dataset represents 1-year of networking data collected from 350 companies with IoT deployments, utilizing the  Arm Mbed IoT Device Management Platform.
The exploratory analysis of the underlying dataset reveals a set of interesting insights into the nature of such IoT deployments. It shows that the IoT environments exhibit properties similar to the earlier studied web and media sites~\cite{arlitt1996, arlitt2001, cherkasova2004, tang2007} and could be described by famous Pareto principle~\cite{pareto}, when the data distribution follows the power law~\cite{power-law}. The Pareto principle (also known as the 80/20 rule or the "law of the vital few") states that for many events or data distributions roughly 80\% of the effects come from 20\% of the causes. For example, in the earlier web sites, 20\% of the web pages were responsible for 80\% of all the users accesses~\cite{arlitt1996}. The later, popular web sites follow a slightly different proportion rule: they often are described by  90/10 or 90/5 distributions, i.e., 90\% of all the user accesses are targeting a  small subset of popular web pages, which represent  5\% or 10\% of the entire web pages set. 

The interesting findings from the studied IoT networking dataset  can be summarized as follows:
\begin{itemize}
\vspace{-2mm}
\item 10\% of the companies in the dataset contribute to 90\% of  the entire traffic;
\item 7\% of all the companies in the dataset own 90\% of all the devices.
\vspace{-2mm}
\end{itemize}
 {\it IoTelligent} tool applies machine learning techniques to forecast the companies’ traffic demands over time, visualize traffic trends, identify and cluster devices, detect device anomalies and failures. We designed and evaluated CNN, LSTM, and Convolutional LSTM models for demand forecasting, with a conclusion of the Convolutional LSTM model being the best. To avoid maintaining and upgrading tens (or hundreds) of models (a different model per company), we designed and implemented a novel, scalable approach, where a global demand forecasting model is built using the combined data of all the companies. The accuracy of the designed approach is further improved by normalizing the “contribution” of individual company data in the combined global dataset. To solve the scalability issues with managing the millions of devices, we designed and evaluated a novel technique based on: (i)    autoencoders, which extract the relevant features automatically from the network traffic stream; (ii) DBSCAN clustering to identify the group of devices that exhibit similar behavior, to flag anomalous devices.
The designed management tool paves the way the industry can monitor their IoT assets for presence, functionality, and behavior at scale without the need to develop device-specific models.
\vspace{-1mm}

\vspace{-2mm}
\section{Dataset and the Exploratory Data Analysis} 
The network traffic data was collected from more than 350 companies for a total duration of one year. The traffic data is binned using 15~minute  time window, used for billing purposes. 
\begin{itemize}
\vspace{-2mm}
\item Unix timestamp;
\item Anonymous company ids;
\item Anonymous device ids per company;
\item The direction of the network traffic (to and from the device);
\item Number of bytes transmitted in the 15 minute interval;
\item Number of packets transmitted in the 15 minute interval.
\vspace{-1mm}
\end{itemize}
\vspace{-1mm}
Preliminary analysis was done to find the most impactful and well-established companies. We found that the companies' data that represent two essential metrics, such as the networking traffic amount and number of deployed IoT devices,  both follow the Pareto law.
The main findings from the studied IoT networking dataset  can be summarized as follows:
\begin{itemize}
\item 10\% of the companies in the dataset contribute to 90\% of  the entire traffic;
\item 7\% of all the companies in the dataset own 90\% of all the devices.
\end{itemize}

\begin{figure}
 \vspace{-4mm}
  \includegraphics[width=1\columnwidth]{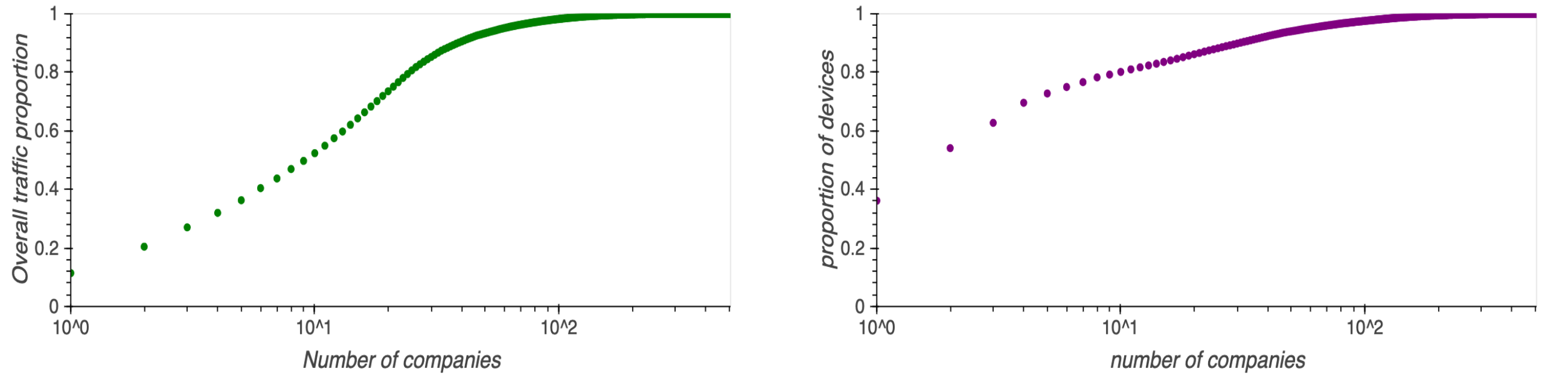}
  \centering
  \vspace{-7mm}
  \caption{\small (left) CDF of networking traffic; (right) CDF of devices.}
  \label{fig:traffic}
  \vspace{+2mm}
 \vspace{+3mm}
  \includegraphics[width=1\columnwidth]{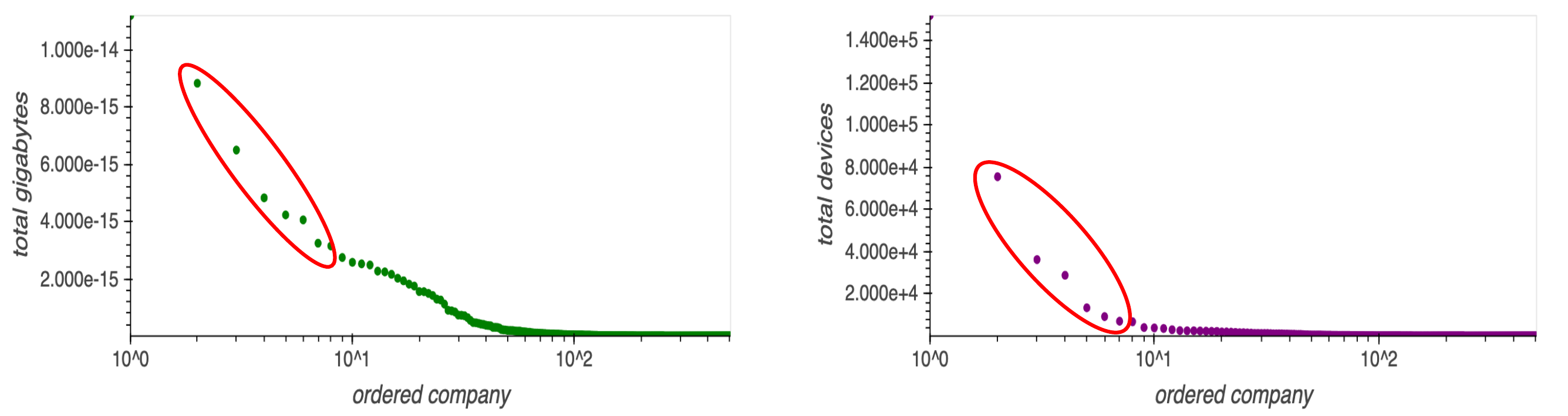}
  \centering
  \vspace{-7mm}
  \caption{\small (left) Networking traffic per company; (right) Number of devices per company.}
  \label{fig:devices}
  \vspace{-4mm}
\end{figure}

Figure~\ref{fig:traffic} shows on the left side the logscale graph of CDF (Cumulative Distribution Function) of the traffic (where one can see that 10\% of the companies in the dataset contribute to 90\% of the entire traffic) and the CDF of the devices per company distribution (where one can see that 7\% of all  the companies in the dataset own 90\% of all the devices). Also, it is quite interesting to note how significant and sizable the contributions are of the first 5-10 companies on those graphs: both for the networking traffic volume and the number of overall devices.

Another interesting observation was that companies with highest number of devices did not correspond to companies with maximum amount of traffic, and vice versa, the high volume traffic companies did not have a lot of devices. This makes sense, for example,  a difference in the outputs of hundreds of simple sensors and a single recording camera.
Among  some other insights into special properties of many  IoT environments (at the networking level)  we observe the pronounced diurnal and weekly patterns,  and changes in the traffic patterns around some seasonal events and holidays. It could be explained by the fact that many IoT environments are related to human and business activities.
\vspace{-4mm}

\section{Demand Forecasting}
The {\it demand forecasting problem} is formulated in the following way. Given a recent month's traffic pattern for a company, what is the expected traffic for this company a week ahead?
This problem requires that a predictive model forecasts the total number of bytes for each hour over the next seven days.  Technically, this framing of the problem is referred to as a multi-step time series forecasting problem, given the multiple forecast steps.  Choosing the right time granularity for (i) making the prediction and (ii) data used in the model, is  another important decision for this type of a problem. 


We found that a reasonable trade-off would be to use 1~hour time granularity. 
This  eliminates the small noises in traffic and also ensures that we have a sufficient data to train our models on.
\vspace{-3mm}
\subsection{Modeling Approach}
Based on our exploratory data analysis, we select 33 companies with largest traffic and 5 companies with largest number of devices.
These companies are responsible for 90\% of the networking traffic volume and 90\% of IoT devices.
Therefore, by designing and evaluating the modeling approach for these companies, we could efficiently cover the demand forecasting for 90\% of the traffic volume and assessing the monitoring solution for 90\% of devices. 

The specific goal is to predict the company traffic for a next week given the previous three weeks of traffic data in an hourly time granularity. We use {\bf a deep learning} based approach for demand forecasting, because deep learning methods are robust to noise, highly scalable, and generalizable. 
We have considered three different deep learning architectures for demand forecasting: CNN, LSTM, and Convolutional LSTM in order to compare their outcome and accuracy.

\noindent{\textbf{Convolutional Neural Network (CNN)~\cite{krizhevsky2012imagenet}:} }
It is a biologically inspired variant of a fully connected layer, which is designed to use minimal amounts of preprocessing. CNNs are made of Convolutional layers that exploit spatially-local correlation by enforcing a local connectivity pattern between neurons of adjacent layers. The main operations in Convolution layers are Convolution, Activation (ReLU), Batch normalization, and Pooling or Sub-Sampling. The CNN architecture, used in our experiments, has 4 main layers. The first three layers are one-dimensional convolutional layers, each with 64 filters and relu activation function, that operate over the 1D traffic sequence. Each convolutional layer is followed by a max-pooling layer of size 2, whose job is to distill the output of the convolutional layer to the most salient elements.  A flatten layer is used after the convolutional layers to reduce the feature maps to a single one-dimensional vector. The final layer is a dense fully connected layer with 168 neurons (24 hours x 7 days) with linear activation and that produces the forecast by interpreting the features extracted by the convolutional part of the model.

\noindent{\textbf{Long Short Term Memory (LSTM)~\cite{gers1999learning,lstm-orig}:}}
It is a type of Recurrent Neural Network (RNN), which takes current inputs and remembers what it has perceived previously in time. An LSTM layer has a chain-like structure of repeating units and each unit is composed of a cell, an input gate, an output gate, and a forget gate, working together. It is well-suited to classify, process, and predict time series with time lags of unknown size and duration between important events. Because LSTMs can remember values over arbitrary intervals, they usually have an advantage over alternative RNNs, Hidden Markov models, and other sequence learning methods in numerous applications. The model architecture, used in our experiments, consists of two stacked LSTM layers, each with 32 LSTM cells, followed by a dense layer with 168 neurons to generate the forecast. 

\noindent{\textbf{Convolutional LSTM~\cite{xingjian2015convolutional}:}}
Convolutional LSTM is a hybrid deep learning architecture that consists of both convolutional and LSTM layers. The first two layers are the one-dimensional Convolutional layers that help in capturing the high-level features from the input sequence of traffic data. Each convolutional layer is followed by a max-pooling layer to reduce the sequence length. They are followed by two LSTM layers, that help in tracking the temporal information from the sequential features, captured by the convolutional layers. The final layer is a dense fully connected layer, that gives a forecasting output. 

We use batch-normalization and dropout layers in all our models to reduce over-fitting and to improve the training speed of our models.

To evaluate the prediction accuracy of the designed models,  we compare the predicted value $X_n^{pred}$  with  the true, measured value $X_n$ using following metrics:
\indent{\textbf{Mean Absolute Error (MAE)}}: 
{\small $$MAE  = \frac{1}{N}\Sigma_{n=1}^{N} { {|X_n - X_n^{pred}|}}$$}
\indent{\textbf{Mean Squared Error (MSE)}}: 
{\small $$MSE = \frac{1}{N}\Sigma_{n=1}^{N}{(X_n -  X_n^{pred})}^2$$}
\vspace{-5mm}
\subsection{Individual Model Per Company}
\begin{figure}
  \includegraphics[width=0.88\columnwidth]{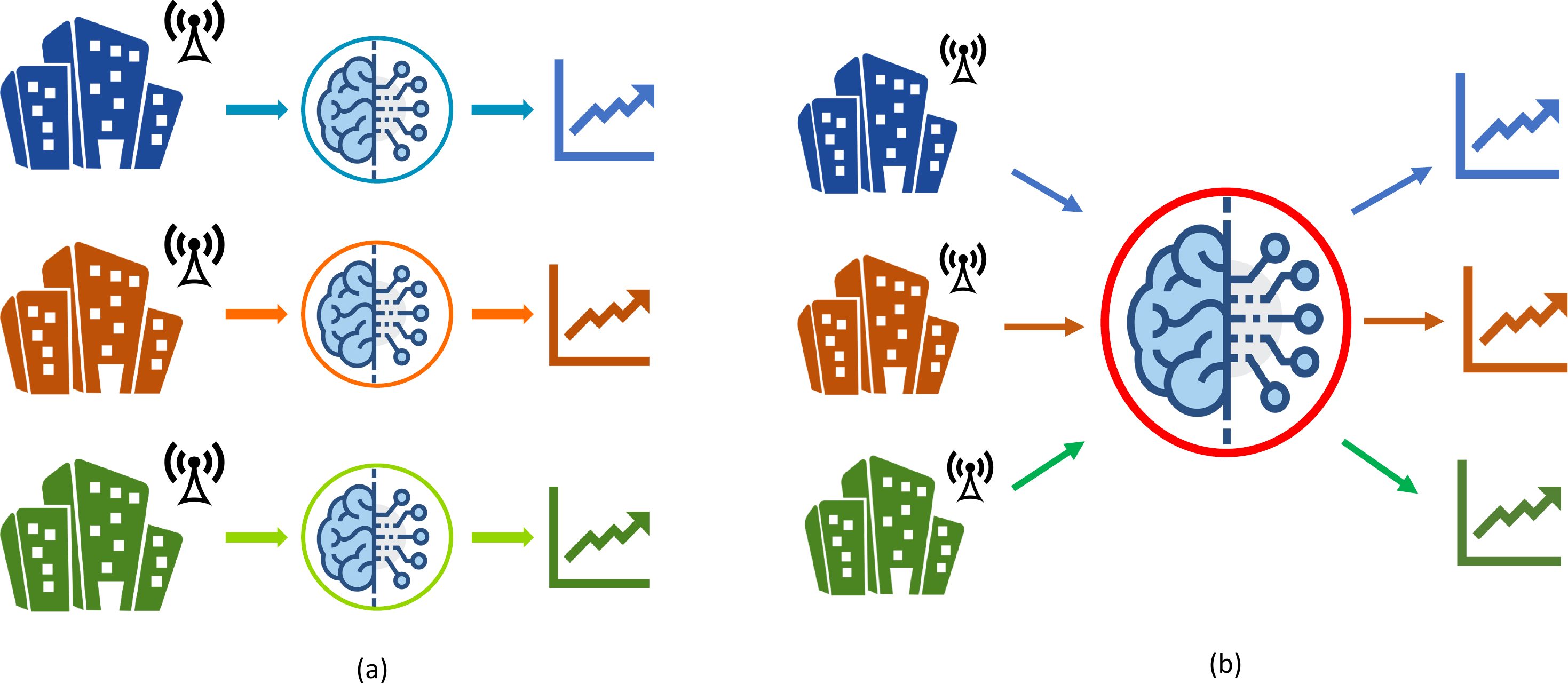}
  \centering
  \caption{\scriptsize (a) Each company has its own prediction model, (b) Using one model for all the companies, trained on the combined dataset.}
  \label{fig:timing-ind}
  \vspace{-4mm}
\end{figure}

This is the naive approach where each company has it's own demand forecasting model, that is, the model for each company is trained by using only the data from that particular company as shown in Figure~\ref{fig:timing-ind}~(a). 
\medskip
\begin{figure}
 \vspace{-6mm}
  \includegraphics[width=0.89\columnwidth]{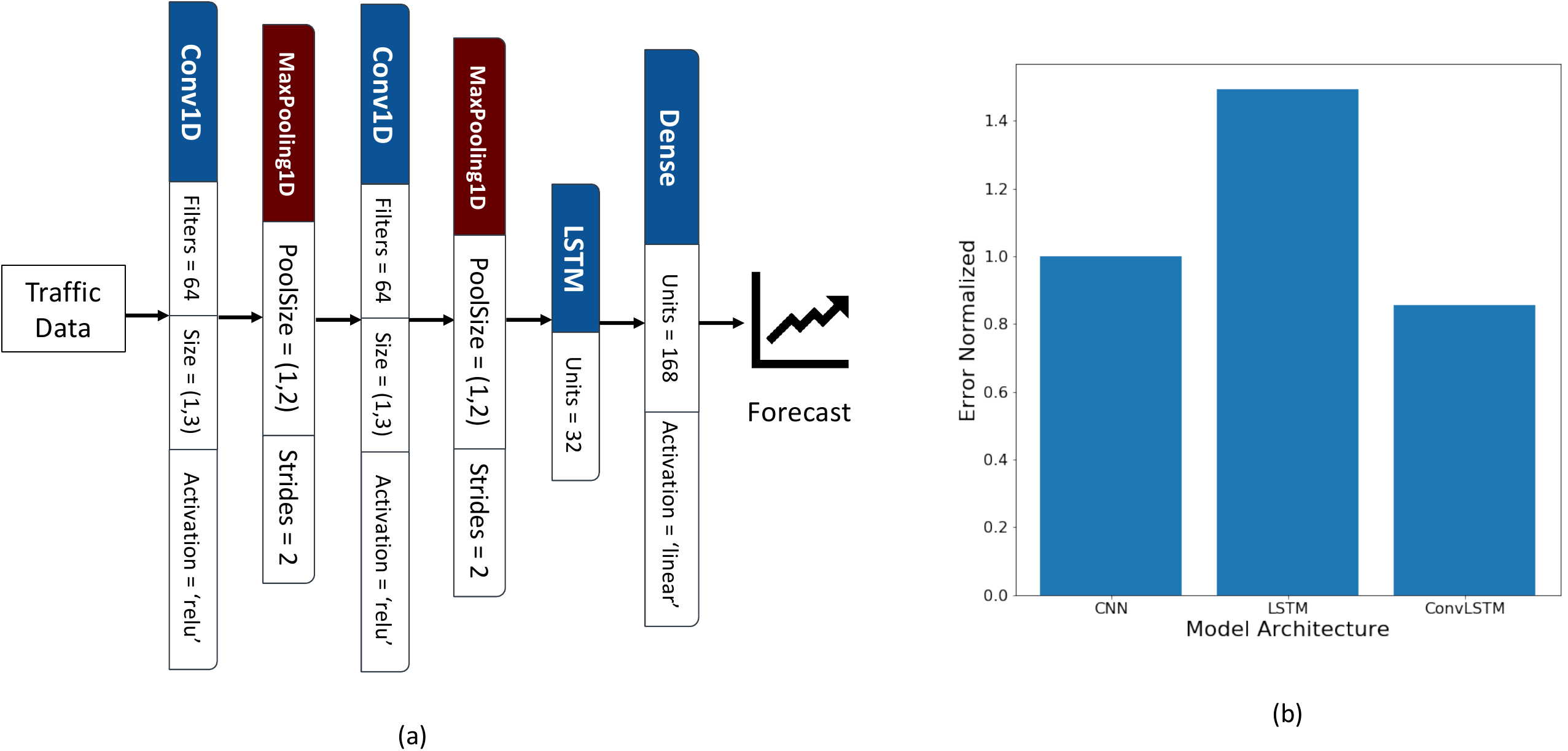}
  \centering
   \vspace{-4mm}
  \caption{\scriptsize (a) Model Architecture of Convolutional LSTM Model; (b) Comparing performance of the three architectures: Convolutional LSTM achieves best performance.}
  \label{fig:relative}
  \vspace{-4mm}
\end{figure}
\begin{figure}
 \vspace{-4mm}
  \includegraphics[scale=0.66]{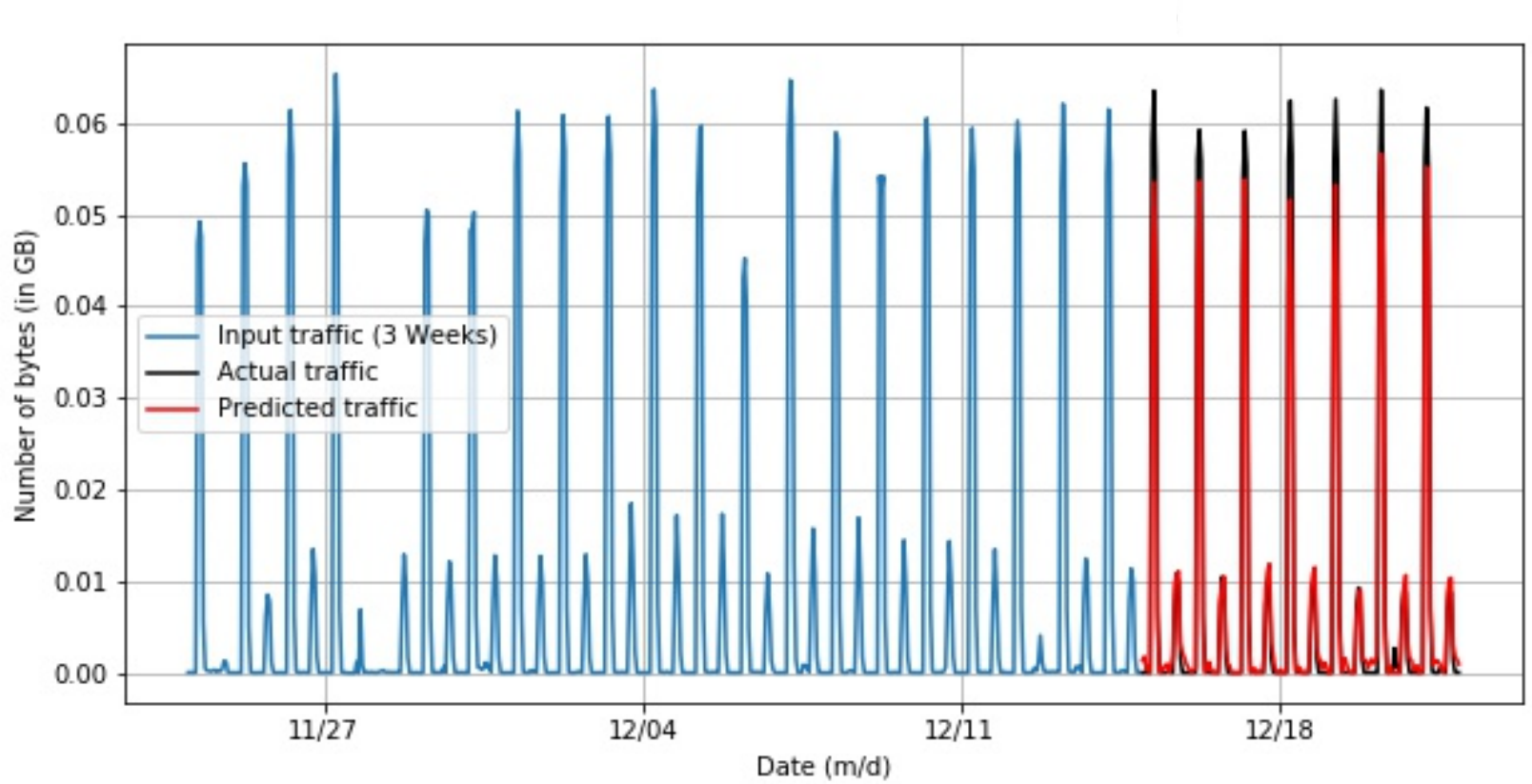}
  \centering
   \vspace{-4mm}
  \caption{\scriptsize Company A: the 4th week demand forecast based on data from the previous 3 weeks.}
  \label{fig:individual}
  \vspace{-4mm}
\end{figure}

So, for each company, we trained three models with the architectures described above (i.e., CNN, LSTM, and Convolutional LSTM). 
Figure~\ref{fig:relative}~(a) presents the detailed parameters of the designed Convolutional LSTM, while 
Figure~\ref{fig:relative}~(b) reflects the relative performance of three different architectures (with the MAE error metrics). 
We found that for both error metrics the Convolutional LSTM model performs better than the other two architectures. 
When comparing architectures accuracy by using MAE and MSE, we can see that Convolutional  LSTM outperforms CNN by {\bf 16\%} and {\bf 23\%} respectively, and outperforms  LSTM by {\bf 43\%} and {\bf 36\%} respectively.
Therefore, only {\bf Convolutional LSTM} architecture is considered for the rest of the paper.
Finally, Figure~\ref{fig:individual} shows an example of company A (in the studied dataset): its measured networking traffic over time  and the forecasting results with the Convolutional LSTM model.

Building an individual model per each company has a benefit that this approach is simple to implement.
However, it is not scalable as the number of required forecasting models is directly proportional to the number of companies. The service provider has to deal with the models' maintenance, their upgrades, and retraining (with new data) over time. 

Therefore, in the next Section~\ref{sec:global-no-norm}, we aim to explore a different approach, which enables a service provider to use all the collected, global data for building a single (global) model, while using it for individual company demand forecasting. 
Only Convolutional LSTM architecture is considered in the remaining of the paper (since as shown, it supports the best performance). 
\vspace{-3mm}

\subsection{One Model for All Companies - Without Normalization}
\label{sec:global-no-norm}

In this approach, we train a single Convolutional LSTM model for demand forecasting by using data from all the companies. The networking traffic data from all the companies were combined. The data from January to October were used for training the model, and the data from November and December were used as the test set. 

\begin{figure}[htb]
 \vspace{-6mm}
  \includegraphics[scale=0.68]{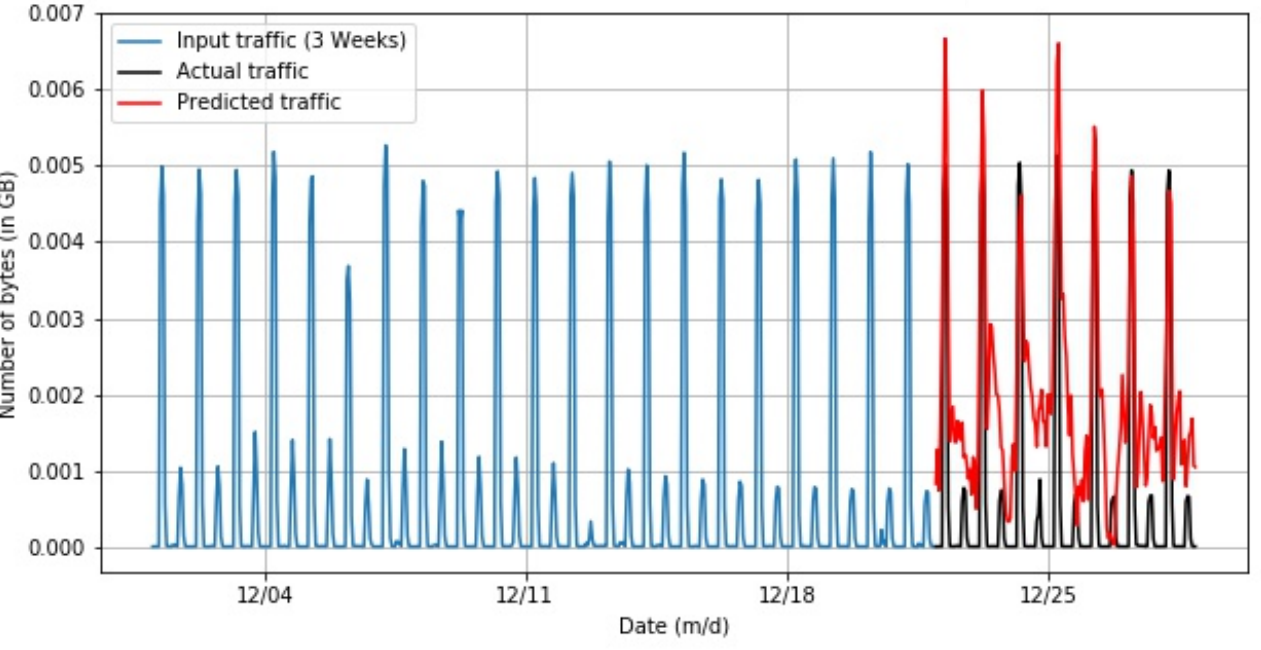}
  \centering
   \vspace{-4mm}
  \caption{\scriptsize Demand forecasting using the Global model trained on data without normalization.}
  \label{fig:plotglobalno}
  \vspace{-5mm}
\end{figure}

This method is highly scalable since it trains and utilizes a single model for demand forecasting of all companies. While this approach is very attractive and logical, it did not always produce good forecasting results. 
Figure~\ref{fig:plotglobalno} shows the forecasting made by this global model for Company A (with this company we are already familiar from Figure~\ref{fig:individual}). As we can see in Figure~\ref{fig:plotglobalno}, the model fails to capture a well-established traffic pattern. 

One of the explanations of the observed issue is that this company traffic constitutes a very small fraction compared to other companies in the combined dataset. So, the globally trained model has  ``learned" the traffic patterns of larger companies in the set, while ``downplayed" the traffic patterns of a smaller company.
 The model fails to capture a well-established traffic pattern for companies with less traffic because the prediction loss in terms of absolute value is still small. However,  it is not a desirable outcome as we would like our model to capture the traffic pattern even for companies with low traffic.

This global model has a major drawback: the model learns to give more importance to companies with high traffic volume than the companies with low traffic. As a result, the network traffic patterns of the companies with small, but well-established traffic patterns, are not captured properly.

\subsection{One Model for All Companies - With Normalization}\label{sec:globalnorm}
\begin{figure}[htb]
  \vspace{-4mm}
  \includegraphics[width=0.67\columnwidth]{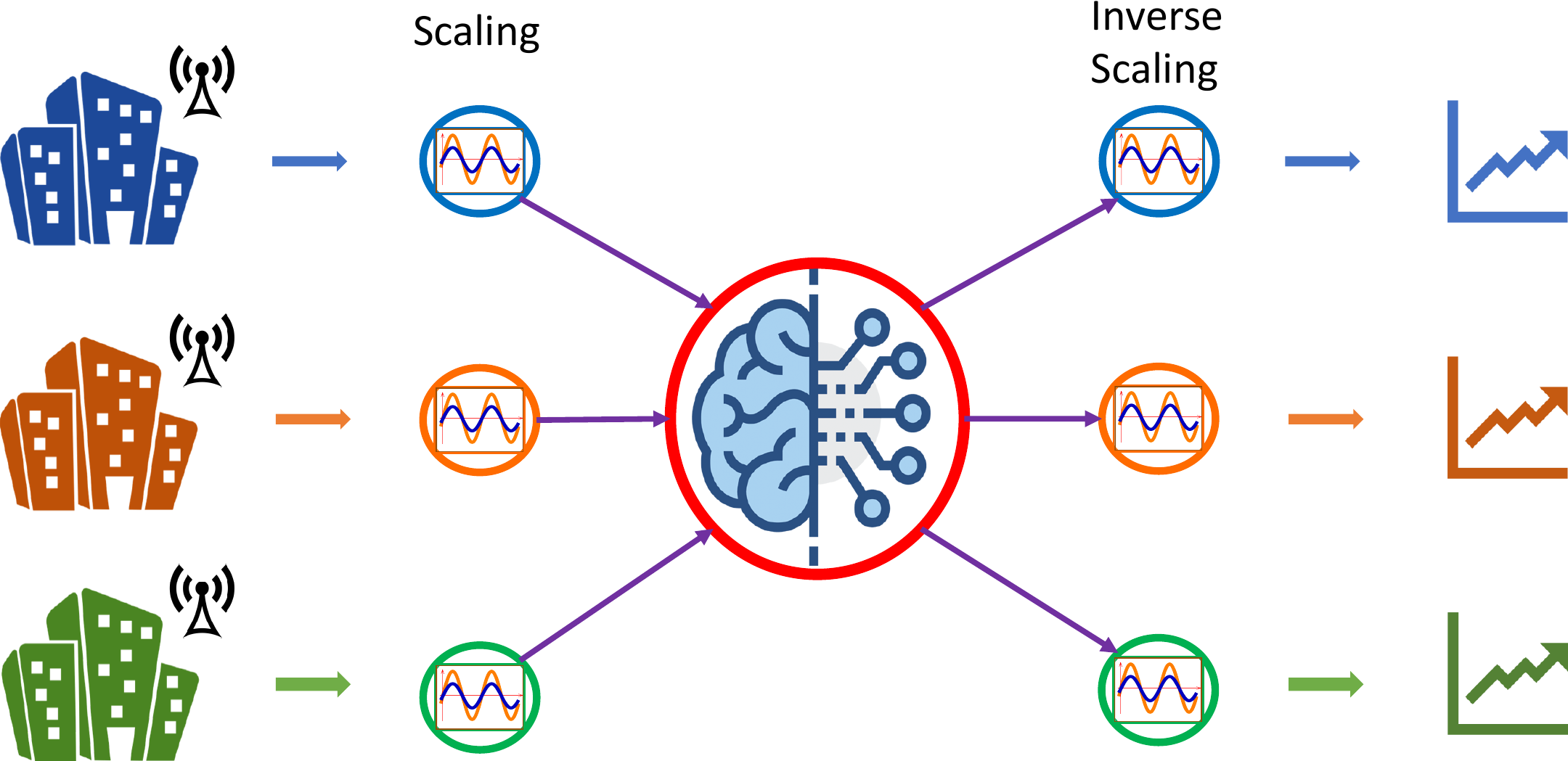}
  \centering
   \vspace{-2 mm}
  \caption{\scriptsize One global prediction model is trained by using the normalized data from all the companies.}
  \label{fig:timing}
  \vspace{-4mm}
\end{figure}

This method aims to address the issues of the previous two approaches. In this method, the data from each company is normalized, that is, all the data subsets are scaled so that they lie within the same range. We use the min-max scaling approach to normalize the data subsets so that the values of the data for all companies lie between 0 and 1. Equation~\ref{eq:1} shows the formula used for min-max scaling, where 'i' refers to the 'i'th company. 
{\small
\begin{equation}\label{eq:1}
X_{norm}^{i} = \frac{X^{i} - X_{min}^{i}}{X_{max}^{i} - X_{min}^{i}}
\end{equation}
}
Then a single deep learning model for forecasting is trained using the normalized data of all companies. The predicted demand (forecast) is then re-scaled using Equation~\ref{eq:2} to the original scale. 
{\small
\begin{equation}\label{eq:2}
X^{i} = X_{norm}^{i} * (X_{max}^{i} - X_{min}^{i}) + X_{min}^{i}
\end{equation}
}
This method of training the global model gives equal importance to the data from all companies and treats them fairly. The designed model does not over-fit and is generalizable since it is trained on the data from multiple companies. Figure~\ref{fig:timing} graphically reflects the process of the global model creation with normalized data from different companies.

\begin{figure}[htb]
  \vspace{-4mm}
  \includegraphics[scale=0.7]{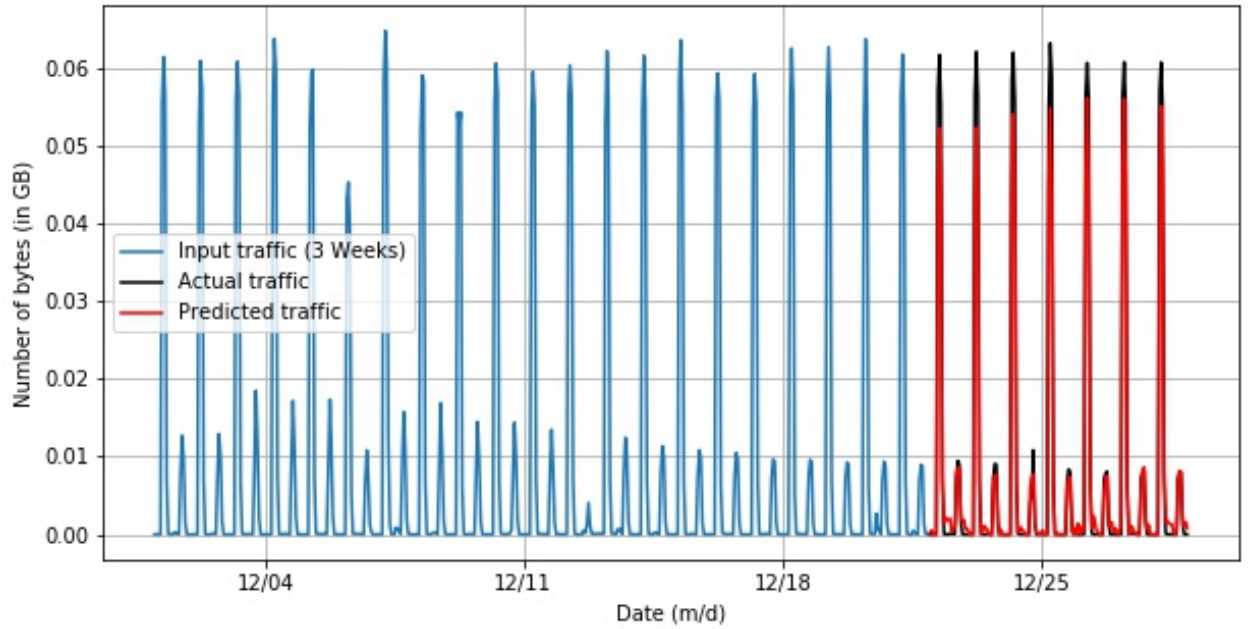}
  \centering
   \vspace{-4mm}
  \caption{\scriptsize Demand forecasting using the Global model trained on data with normalization.}
  \label{fig:plotglobalyes}
  \vspace{-4mm}
\end{figure}

Figure~\ref{fig:plotglobalyes} shows that the designed forecasting model can capture well the patterns of companies with low traffic volume (such as Company~A).

\section{Introducing Uncertainty to Forecasting Models}\label{sec:uncertainty}
\begin{figure}
  \vspace{-8mm}
  \includegraphics[scale=0.34]{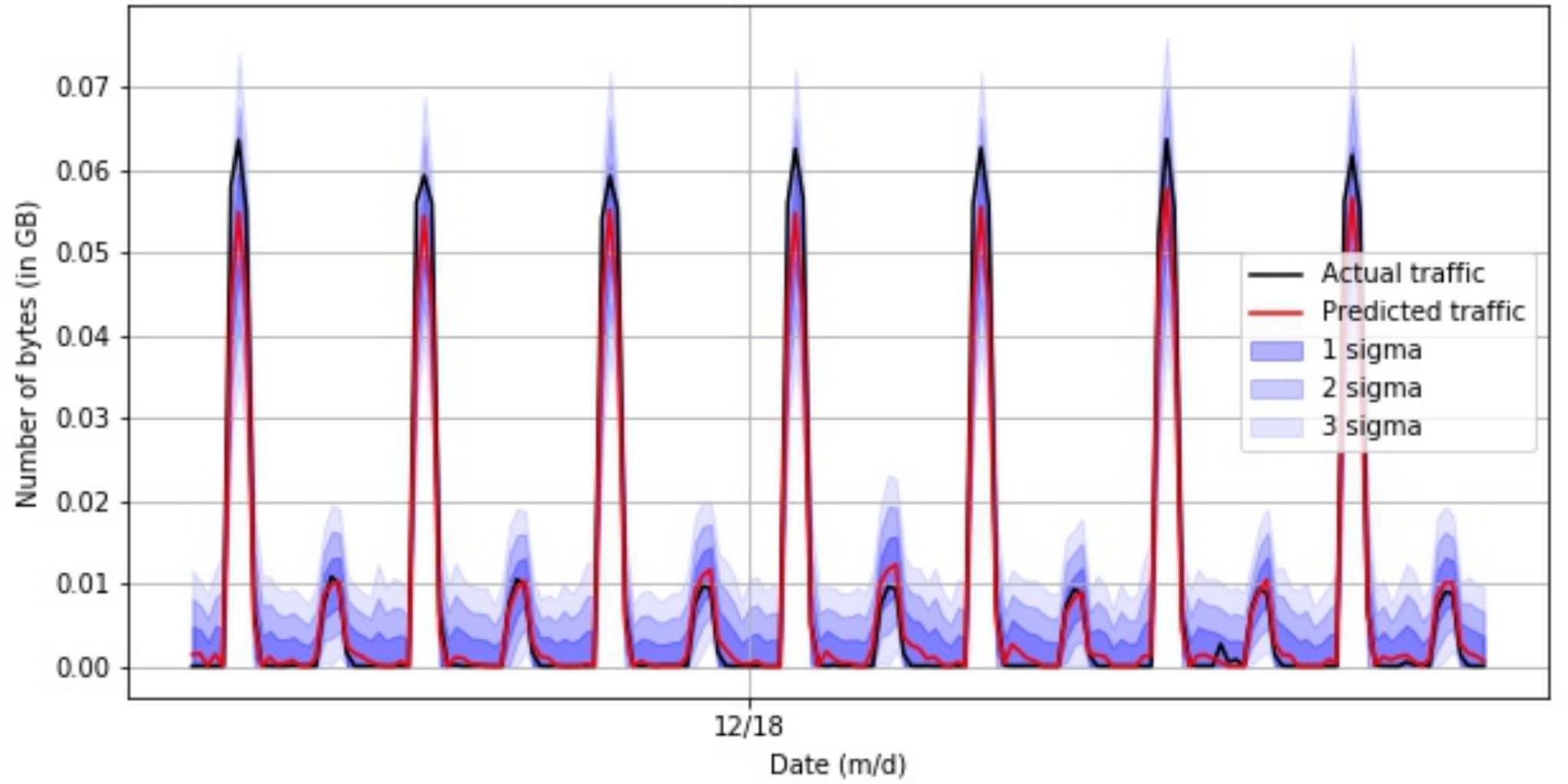}
  \centering
    \vspace{-4mm}
  \caption{\scriptsize Demand forecasting with uncertainty for Global model trained on data with normalization.}
  \label{fig:forecastunc}
  \vspace{-8mm}
\end{figure}

In the previous section, we designed a single global model with normalization, that can be used to forecast for multiple companies. But demand forecasting is a field, where an element of uncertainty exists in all the predictions, and therefore, representing model uncertainty is of crucial importance. The standard deep learning tools for forecasting do not capture model uncertainty. Gal et. al ~\cite{gal2016dropout} propose a simple approach to quantify the neural network uncertainty, which shows that the use of dropout in neural networks can be interpreted as a Bayesian approximation of a Gaussian process -  a well known probabilistic model. Dropout is used in many models in deep learning as a way to avoid over-fitting, and it acts as a regularizer. However, by leaving it ``on" during the prediction, we end up with the equivalent of an ensemble of subnetworks, within our single larger network, that have slightly different views of the data. If we create a set of $T$ predictions from our model, we can use the mean and variance of these predictions to estimate the prediction set uncertainty. Figure~\ref{fig:forecastunc} shows the forecast with uncertainty for Company A, using the global model with normalization. 
\medskip

To evaluate the quality of forecast based on uncertainty, we introduce Prediction Interval Coverage Probability (PICP) metric. 

\subsection{Prediction Interval Coverage Probability (PICP)}
PICP tells us the percentage of time an interval contains the actual value of the prediction. Equations 3-5 show the calculation of PICP metric, where $l$ is the lower bound, $u$ is the upper bound, x\textsubscript{i} is the value at timestep $i$, \^{y} is the mean of the predicted distribution, $z$ is the number of standard deviations from the Gaussian distribution, (e.g., 1.96 for a 95\% interval), and $\sigma$ is the standard deviation of the predicted distribution.
{\small}
\begin{equation}
l(x_{i}) = \hat{y}_{i} - z * \sigma_{i}
\end{equation}
\begin{equation}
u(x_{i}) = \hat{y}_{i} + z * \sigma_{i}
\end{equation}
\begin{equation}
PICP_{l(x),u(x)} = \frac{1}{N}\sum_{i=1}^{N}h_{i}, \quad where \; h_{i}=
\left\{\begin{matrix}
1, & if \; l(x_{i})\leq y_{i}\leq u(x_{i})\\ 
0, & otherwise
\end{matrix}\right.
\end{equation}

\subsection{Evaluating Forecast with Uncertainty}
\begin{figure}
  \includegraphics[width=0.92\columnwidth]{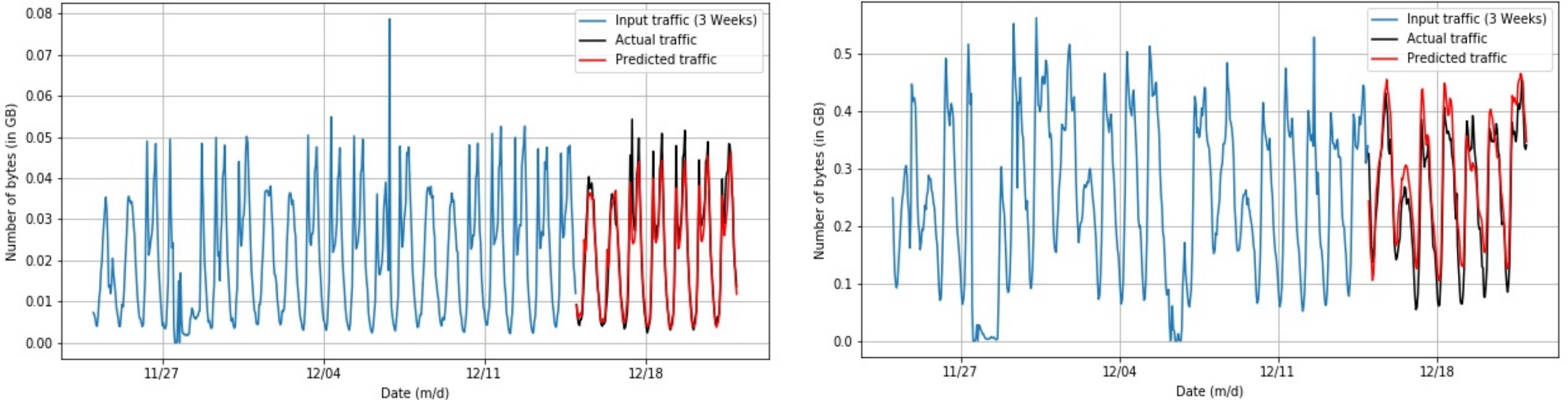}
  \centering
  \caption{\small Demand forecasting with Global model trained on data with normalization.}
  \label{fig:forecast}
  \vspace{-4mm}
\end{figure}

\begin{figure}
  \includegraphics[width=0.92\columnwidth]{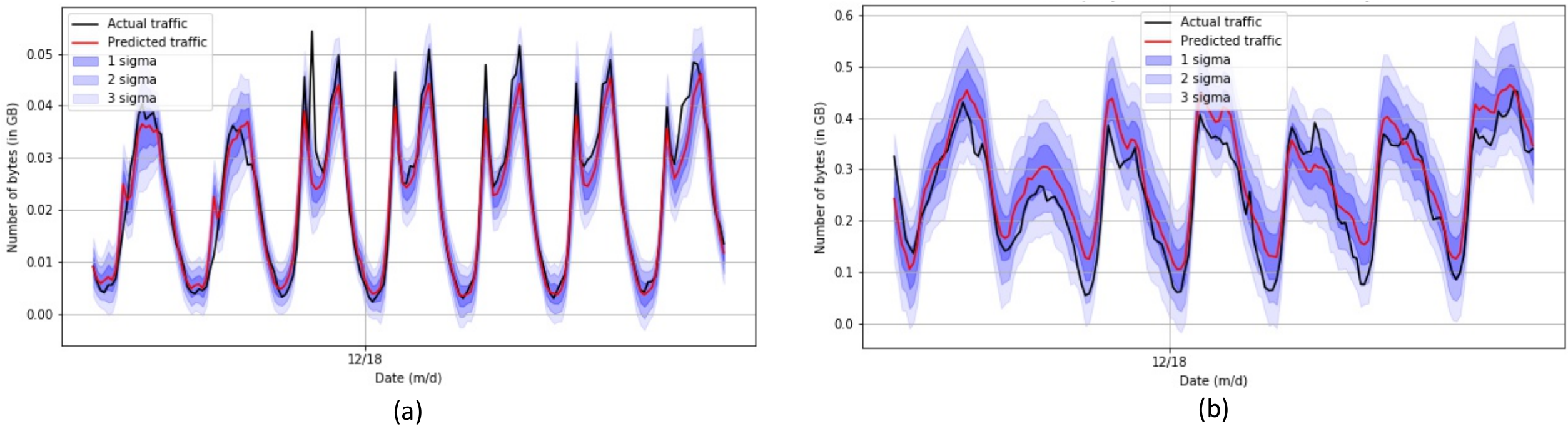}
  \centering
  \caption{\small Demand forecasting with Uncertainty using the Global model trained on data with normalization.}
  \label{fig:uncertainty}
  \vspace{-4mm}
\end{figure}

\begin{figure}
  \vspace{8mm}
  \includegraphics[scale=0.3]{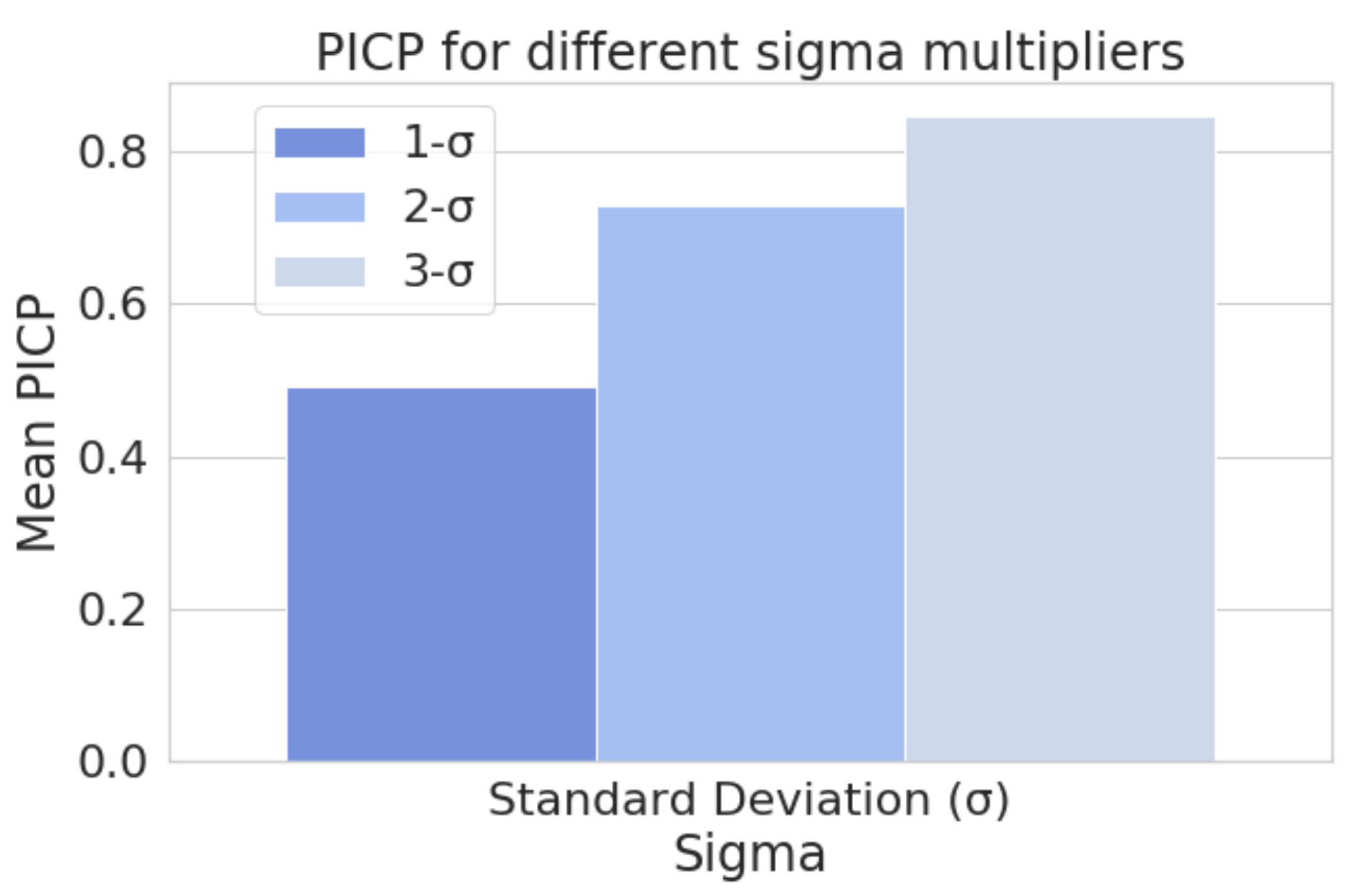}
  \centering
  \caption{Mean PICP for different values of sigma multiplier.}
  \label{fig:PICP}
  \vspace{-4mm}
\end{figure}

We evaluate the overall performance of our global forecast model, introduced in Section~\ref{sec:globalnorm}, based on the PICP metric described above. The forecasting is done 100 times for each company with a dropout probability of 0.2, and then the mean and standard deviations are obtained for each company. Figure~\ref{fig:forecast} shows the global model's forecast for the third week of December for two different companies: Company B and Company C. As we can see from the plot, the model captures the traffic pattern, but still, the predicted values show some deviations from the actual values. This results in some errors when using the traditional error metrics discussed in Section~\ref{sec:error-metrics}, though the model is performing very well. Therefore, introducing uncertainty helps the model to generate a reasonable forecast distribution. Figure~\ref{fig:uncertainty} shows the forecast with uncertainty, where the different shades of blue indicate the uncertainty interval, obtained for different values of uncertainty multipliers. As we can see from the plot, the single global forecasting model can capture well the general traffic trends across multiple companies. Figure~\ref{fig:PICP} shows the mean PICP calculated across all the companies for the different uncertainty multipliers.

We find that on an average 50\% of the forecast values lie within the predicted interval with one standard deviation, 74\% for two standard deviations and 85\% for three standard deviations. The forecast samples which lied outside the predicted interval were mostly due to the fact that the months of November and December had lots of holidays and hence those days did not follow the captured traffic pattern. 
\section{Device Monitoring and Diagnostics}
Once an Internet of Things (IoT) ecosystem is installed, it does not follow a “fire and forget” scenario. There will be unforeseen operational issues, some devices will fail and/or would need to be either repaired or replaced. Each time this happens, the company is on the mission to minimize the downtime and ensure that its devices function properly to protect their revenue stream. However, to address the issues of failed or misbehaving devices, we need to identify such devices in the first place. Therefore, the ability to monitor the device's health and being able to detect, when something is amiss, such as higher-than-normal network traffic or ``unusual" device behavior, it is essential to proactively identify and diagnose potential bugs/issues. Again, large-scale IoT deployments is a critical and challenging issue. When there are thousands of devices in the IoT ecosystem, it becomes extremely difficult to efficiently manage these devices as it is practically impossible to monitor each device individually. So, we need an efficient way to analyze the observed device behaviors and identify devices that show an anomalous (``out of usual norm") behavior. 

Anomalous or failed devices can be categorized into two types:
\begin{enumerate}
    \item The devices that behave significantly different from the other devices;
    \item The devices whose observed behavior suddenly changes from its ``normal" behavior over time.
\end{enumerate}
The following Section describes the designed technique to accomplish the device monitoring and diagnostic via device categorization over time. 

\subsection{Clustering to identify Anomalies and group Devices based on their traffic patterns}
\label{sec:clustering}

\begin{figure}
  \includegraphics[width=1\columnwidth]{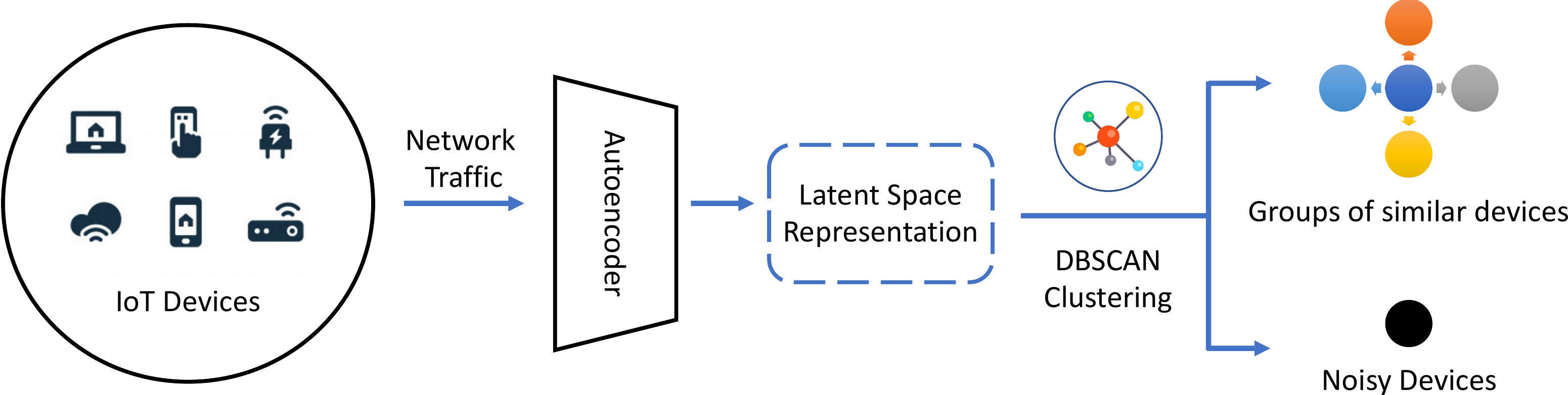}
  \centering
  \caption{Identifying similar groups of devices.}
  \label{fig:device management}
  \vspace{-4mm}
\end{figure}

When there are thousands of devices in a given IOT Ecosytem, there usually exist multiple devices of the same type or having similar behavior. We identify these groups of devices in an unsupervised manner based on their network traffic pattern over a given month. Figure ~\ref{fig:device management} shows an overview of the proposed method and its steps to obtain the groups of "similar" devices:
\begin{itemize}
    \item The monthly network traffic from the thousands of IoT devices are passed through an autoencoder to extract features in the latent space in an unsupervised manner.
     \item Then we use a density-based clustering algorithm, DBSCAN, on the latent space to identify the groups of similar devices.  The objective is to learn what normal data points looks like and then use that to detect abnormal instances. Any instance that has a low affinity to all the clusters is likely to be an anomaly.
\end{itemize}

\subsubsection{Autoencoder~\cite{masci2011stacked}}

It is a neural network capable of learning dense representations of the input data, called latent space representations, in an unsupervised manner. The latent space has low dimensions which helps in visualization and dimensionality reduction. An autoencoder has two parts: an encoder network that encodes the input values x, using an encoder function f, and, a decoder network that decodes the encoded values f(x), using a decoder function g, to create output values identical to the input values. Autoencoder‘s objective is to minimize reconstruction error between the input and output. This helps autoencoders to capture the important features and patterns present in the data in a low dimensional space. When a representation allows a good reconstruction of its input, then it has retained much of the information present in the input. In our experiment, an autoencoder is trained using the monthly traffic data from the IoT devices which captures the important features or the encoding of the devices in the latent space.

\noindent \textbf{Architecture of the Autoencoder:}
We use a stacked autoencoder in our experiment with two fully connected hidden layers each in the encoder and the decoder. The central bottle neck layer was a fully connected layer with just three neurons which helps in reducing the dimensions. We used mean squared error as the reconstruction loss function.

\subsubsection{DBSCAN~\cite{ester1996density}}
(Density-Based Spatial Clustering of Applications with Noise), is a density-based clustering algorithm that captures the insight that clusters are dense groups of points. If a particular point belongs to a cluster, it should be near to lots of other points in that cluster. The algorithm works in the following order: First, we choose two parameters, a positive number, epsilon and a natural number, minPoints. We then begin by picking an arbitrary point in our dataset. If there are more than minPoints points within a distance of epsilon from that point, (including the original point itself), we consider all of them to be part of a "cluster". We then expand that cluster by checking all of the new points and seeing if they too have more than minPoints points within a distance of epsilon, growing the cluster recursively if so. Eventually, we run out of points to add to the cluster. We then pick a new arbitrary point and repeat the process. Now, it's entirely possible that a point we pick has fewer than minPoints points in its epsilon ball, and is also not a part of any other cluster. If that is the case, it's considered a "noise point" not belonging to any cluster and we mark that as an anomaly.
\begin{figure}
\vspace{-8mm}
  \includegraphics[width=1\columnwidth]{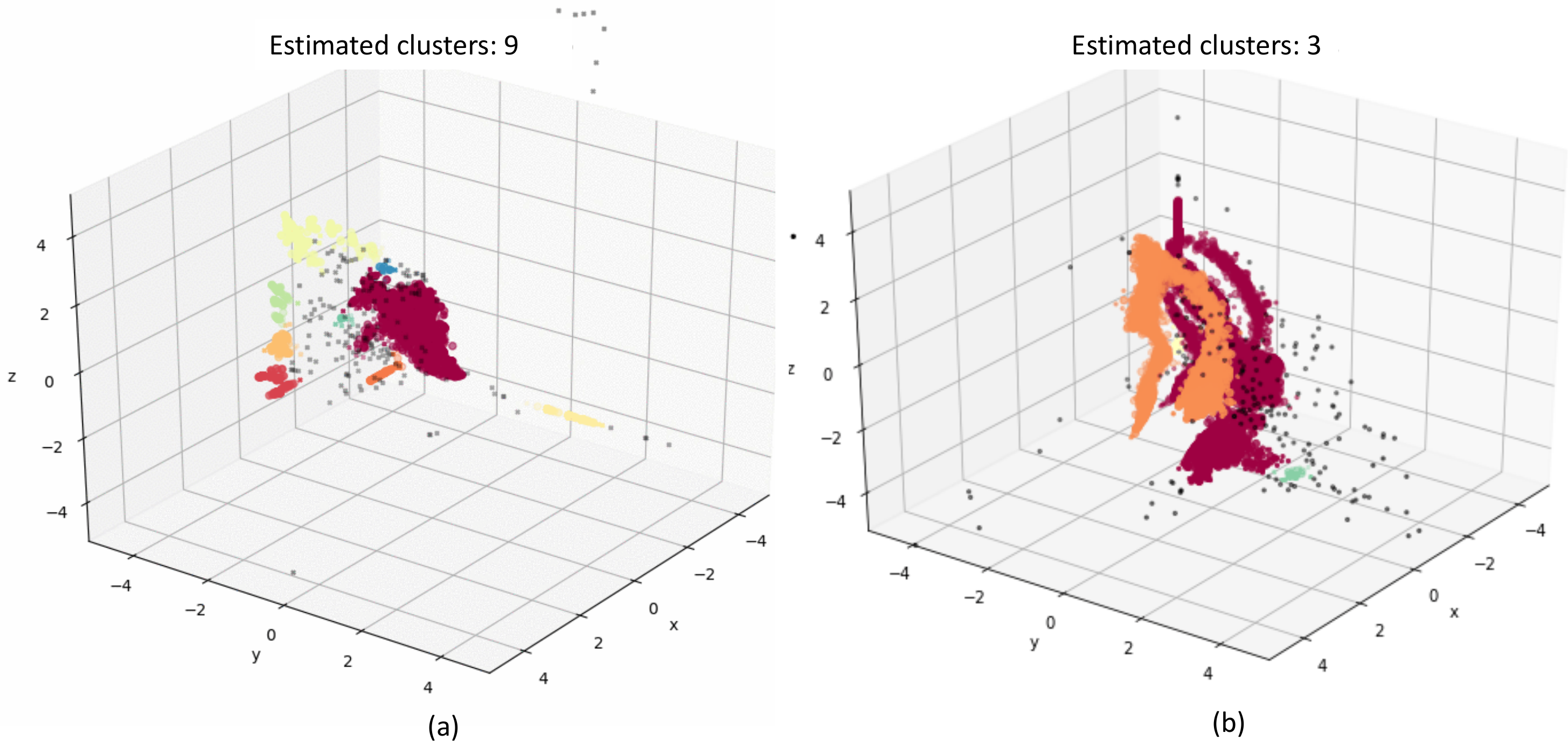}
  \centering
  \caption{Number of Clusters per company when visualized in the latent space}
  \label{fig:clusters}
  \vspace{-8mm}
\end{figure}

Figure~\ref{fig:clusters} shows the latent space and the clusters obtained for Company A(left) and Company B(right). Companies A and B had more than 30000 devices each, installed in their IoT ecosystems and they had three and nine unique types of devices respectively. Based on their traffic patterns observed over the period of a month, the autoencoder mapped the devices of the same type close to each other while the devices of different types were mapped far apart from each other in the latent space. When DBSCAN clustering was applied in the latent space, we observed that the number of distinct clusters formed was exactly the same as the corresponding number of device types per company. The devices which didn't fall in these well formed clusters because of their different traffic patterns were marked as anomalies and are represented by the black points.

\section{Related Work}
\label{sec:related}

Demand forecasting has been broadly studied due to the problem importance and its significance for utility companies.{\it Statistical methods} use historical data to make the forecast as a function of most significant variables.  The detailed survey on regression analysis for the prediction of residential energy consumption is offered in~\cite{regr-survey}. The authors believe that among statistical models, linear regression analysis has shown promising results because of satisfactory accuracy and simpler implementation compared to other methods. In many cases, the choice of the framework and the modeling efforts are driven by the specifics of the problem formulation. 

While different studies have shown that demand forecasting depends on multiple factors and hence can be used in multivariate modeling, the univariate methods like ARMA and ARIMA~\cite{ARMA2002,ARIMA2017} might be sufficient for short term forecast. {\it Machine learning} (ML) and {\it artificial intelligence} (AI) methods based on neural networks~\cite{NN2016,NN2017}, support vector machines SVM)~\cite{SVM1}, and fuzzy logic~\cite{fuzzy1} were applied to capture complex non-linear relationships between inputs and outputs. When comparing ARIMA, traditional machine learning, and artificial neural networks (ANN)  modeling, some recent articles provide contradictory results. In~\cite{arima-vs-ANN-1}, ARIMA achieves better results than ANN, while the study~\cite{arima-vs-ANN-2} claims that ANNs perform slightly better than ARIMA methods. 
In our work,  we construct a deep-learning based Convolutional LSTM forecasting model (a hybrid model with both Convolutional and LSTM layers). The Convolutional LSTM model works well for long term (weekly) demand prediction, and indeed, automatically captures non-linear daily patterns. 

In general, the quality and the prediction power of the models designed by using ML and AI methods critically depend on the quality and quantity of historical data. To create a good forecasting model, several approaches have been developed in the literature. One such approach is an ensemble of multiple forecasting methods applied on the same time series data and a weighted average of their forecasts is used as a final result ~\cite{automate-forecast4}. In our work, we pursue a different approach by making use of the normalized data from multiple companies and train a single global model to make traffic predictions. This makes our method highly scalable. 

\section{Conclusion}
In our work, we proposed {\it IoTelligent}, a tool that applies machine learning techniques to forecast the companies’ traffic demands over time, visualize traffic trends, identify and cluster devices, detect device anomalies and failures. We showed that among the different neural network architectures, Convolutional LSTM model performed the best for demand forecasting. In order to avoid maintaining and upgrading tens (or hundreds) of models (a different model per company), we designed and implemented a novel, scalable approach, where a global demand forecasting model is built using the combined data of all the companies. This method was improved by normalizing the “contribution” of individual company data in the combined global dataset. We also introduced uncertainty intervals to the forecasts to provide better information to the users. To solve the scalability issues with managing the millions of devices, we designed and evaluated a novel technique based on: (i)	autoencoders, which extract the relevant features automatically from the network traffic stream; (ii) DBSCAN clustering to identify the group of devices that exhibit similar behavior, in order to flag anomalous devices. The designed management tool  paves the way the industry can monitor their IoT assets for presence, functionality, and behavior at scale without the need to develop device specific models.

{\scriptsize}
%
%

\clearpage
\end{document}